\documentclass[journal]{IEEEtran}
\usepackage{epsfig}
\usepackage{graphicx}
\usepackage{algorithmic}
\usepackage{amsmath}
\usepackage{indentfirst}
\usepackage{cases}
\usepackage{amsmath,bm}

\usepackage{amssymb}
\usepackage{times}
\usepackage{graphicx}
\usepackage{subfigure}
\usepackage{multirow}
\usepackage{multicol}
\usepackage{amsmath}

\usepackage[ruled,vlined,linesnumbered]{algorithm2e}

\usepackage{threeparttable}
\usepackage{dcolumn}
\usepackage{multirow}
\usepackage{multicol}
\usepackage{booktabs}
\usepackage{subfigure}

\usepackage{amsthm}

\usepackage{latexsym}

\usepackage{rotating}
\usepackage{textcomp}

\usepackage{textcomp,booktabs}
\usepackage[usenames,dvipsnames]{color}
\usepackage{colortbl}
\definecolor{mygray}{gray}{.7}
\definecolor{mypink}{rgb}{.99,.91,.95}
\definecolor{mycyan}{cmyk}{.3,0,0,0}

\usepackage{xcolor}
\usepackage[normalem]{ulem} 

\newtheorem{remark}{Remark}[section]

\newcommand\hl{\bgroup\markoverwith
  {\textcolor{yellow}{\rule[-.5ex]{2pt}{2.5ex}}}\ULon}

\begin{document}        
%
\title{Weights Adaptation Optimization of Heterogeneous Epidemic Spreading Networks: A Constrained Cooperative Coevolution Strategy}


\author{\IEEEauthorblockN{Yun Feng and Bing-Chuan Wang}

\thanks{Yun Feng is with the Department of Systems Engineering and Engineering
	Management, City University of Hong Kong, Kowloon, Hong Kong, and also with the Central South University, Changsha, 410083, P. R. China. \protect E-mail: yun.feng@my.cityu.edu.hk
	
	Bing-Chuan Wang is with the School of Automation, Central South University, Changsha, 410083, P. R. China. \protect E-mail: wangbingchuancityu@gmail.com (Corresponding author: BC Wang) }

}

 \maketitle
{\color{blue}
\begin{abstract}
In this paper, the dynamic constrained optimization problem of weights adaptation for heterogeneous epidemic spreading networks is investigated. Due to the powerful ability of searching global optimum, evolutionary algorithms are employed as the optimizers. One major difficulty is that the dimension of the problem is increasing exponentially with the network size and most existing evolutionary algorithms cannot achieve satisfiable performance on large-scale optimization problems. To address this issue, a novel constrained cooperative coevolution ($C^3$) strategy, which can separate the original large-scale problem into different subcomponents, is employed to achieve the trade-off between the constraint and objective function. 
\end{abstract}
}

%
\IEEEpeerreviewmaketitle
\section{Introduction}\label{sec:introduction} 
\IEEEPARstart{E}{pidemic} spreading over complex networks \cite{pastor2015epidemic} has attracted lots of attention since the pioneering work~\cite{bernoulli1760essai} of Daniel Bernoulli in 1760. Many researches have been focused on the mathematical modeling of disease spreading process, classical epidemic models such as the susceptible-infected-susceptible (SIS) \cite{feng2014epidemic,qu2017sis,feng2016CPB,huang2016epidemic}, and the susceptible-infected-recovered (SIR) \cite{nadini2018epidemic} model have been well studied for decades. Since the spreading of disease (such as AIDS, SARS, etc.) may cause numerous damages to the human society, developing control policies for epidemic spreading process is of great significance, with potential applications in public health. As pointed out in \cite{nowzari2016analysis}, the two common strategies to suppress epidemic spreading scale are increasing the recovery rate and decreasing the infection rate. For example, in \cite{ghezzi1997pid}, the PID control laws were implemented for the classical SIR model where the vaccination rate is the control variable. Despite these innovative results, these studies did not take the ``budget'' or the so-called ``control cost'' into consideration, {\color{blue}which must be taken into account in real-world scenarios}. In recent years, optimal control of epidemic spreading \cite{li2017minimizing,nowzari2017optimal,chen2017optimal,pizzuti2018genetic} has gained more and more attentions. For example, in \cite{xu2017optimal}, an optimal control strategy was designed for the vaccination, quarantine and treatment actions. {\color{blue}For inhomogeneous epidemic dynamics, the optimal control problem was studied in \cite{shang2013optimal}.} 

Besides the above control strategies which are focusing on the epidemic spreading parameters, recently another control strategy which aims at adjusting topology of the underlying network was investigated \cite{feng2018CNSNS}. Acting as the ``bridge'' for epidemic spreading from infected individuals to healthy ones, the network topology determines the epidemic transmission efficiency. A quantitative parameter that defines the strength of two nodes in a network is the value of weight between them. Since the physical meaning of the weight can be described by the contact frequency of two individuals, the intuitive idea of controlling the weights is more natural and practical than controlling the spreading parameters. In \cite{hu2018individual},  an individual-based weight adaptation mechanism in which individuals' contact strength is adaptable depending on the level of contagion spreading over the network was proposed. An optimal control formulation was also presented to address the trade-off between the global infected level and the local weight adaptation cost corresponding to the topology of the underlying contact network. In \cite{feng2018CNSNS} and \cite{hu2018individual}, the objective function contains both the infection cost and control cost. However, the problem of minimizing the infection cost with given fixed number of budget, i.e., the control cost, has not been studied, which is more practical than the unconstrained optimization problems. This motivates us to investigate the constrained optimization problem of weights adaptation.   

Meanwhile, when solving the optimal control problems in both \cite{feng2018CNSNS} and \cite{hu2018individual}, the forward-backward sweep method (FBSM) \cite{mcasey2012convergence} was used to find the numerical solutions, which is an indirect method itself. The intuitive idea of FBSM is that the initial value problem of the state equations is solved forward in time, using an estimate for the control and adjoint variables. Then the adjoint final value problem is solved backwards in time. These complicated procedures add several difficulties to the optimal problem, which can be summarized as follows:
\begin{itemize}
	\item Need to compute various partial derivatives of the Hamiltonian and solve additional differential equations, which introduce more errors for the optimization problem. 
	\item Need to make an initial guess of the adjoint variables, the sensitivity of the method to changes in initial guesses.
\end{itemize}
In \cite{mcasey2012convergence}, the authors found that the FBSM method fails to terminate under some circumstances. 

Evolutionary algorithms (EAs) \cite{back1997handbook,vcrepinvsek2013exploration} which are inspired by nature, have shown powerful searching ability for the global optimum with few restrictions. Also, since EAs are direct methods, the implementations are much simpler than other conventional methods. EAs have been widely used in the community of network science and engineering~\cite{gong2018community,du2019network,wu2018particle}. For example, a novel memetic algorithm which can preserve the community structure is proposed for the network embedding problem in \cite{gong2018community}. As one of the most powerful evolutionary algorithm, differential evolution (DE) \cite{das2011differential,wang2011differential,liu2010hybridizing} has shown superior performance over other heuristic algorithms on very complex searching and optimization problems. The control parameters of DE are few and it is highly efficient \cite{das2011differential}. This motivates us to solve the constrained optimization problem based on DE. However, since the weights adaptation involves all links in the network for a given period of time \cite{hu2018individual}, the dimension of the optimal solution is relatively high. Solving this kind of large-scale optimization problem \cite{yang2008large} is a challenging problem in the community of evolutionary computation. In addition, how to achieve the trade-off between the constraint and objective function is another difficulty.        

Motivated by the above considerations, a dynamic constrained optimization problem of weights adaptation for heterogeneous epidemic spreading networks based on SIS model {\color{blue}(In this manuscript, we only consider SIS model and SIR model will be studied in the future work)} is formulated.  To deal with the high-dimensional optimization problem, a novel constrained cooperative coevolution ($C^3$) strategy which can separate the original high-dimensional search space into some low-dimensional ones by random grouping strategy is proposed. The $\epsilon$ constraint-handling technique is employed to achieve the trade-off between the constraint and objective function. Moreover, as a commonly used variant of the classical DE, the differential evolution with neighborhood search (NSDE) \cite{yang2007making} algorithm is employed as the optimizer for these sub-problems. 

The main contributions of this paper can be summarized as follows:
\begin{itemize}
	{\color{blue}
	\item A dynamic constrained optimization problem of weights adaptation for heterogeneous
	epidemic spreading networks is formulated.
	\item Evolutionary computation techniques with strong searching ability as well as quite a small amount of demands are applied to solve this kind of problem.  
	\item A novel constrained cooperative coevolution ($C^3$) strategy is tailored for this real-world large-scale optimization problem. 
	\item It is easy to implement and the optimization process can be done once the basic network and epidemic parameters are given. For the feedback control strategy~\cite{shang2015global}, information about the number of susceptible or infected individuals is needed for the control law updating while the proposed method does not require such information. }
\end{itemize}

{\color{blue}The proposed strategy has strong potentials to be applied to real-world scenarios for disease control. Actually the most widely and successfully used strategy in real-world, quarantine, is a special form of weights adaptation. In addition, the proposed strategy has no restrictions on the spreading network.}

The rest of this paper is organized as follows. The model description and problem formulation are given in Section \ref{section:Model}. Differential evolution is briefly introduced in Section \ref{section:DE and FR}. In Section \ref{section:methodology}, the methodology is given in detail. Some numerical experiments are presented in Section \ref{section:Numerical Experiments}. Finally, this paper is concluded in Section \ref{section:conclusion}.

\section{Model Description and Problem formulation} 
\label{section:Model}
\subsection{Heterogeneous Weighted SIS-based Network Model}
In the heterogeneous weighted SIS-based network, all nodes can be classified into two possible states according to their health status, that is, susceptible and infected. Susceptible individuals can be infected by infected individuals through the links between them and in state $X_i(t)=1$. Meanwhile, infected individuals can be cured and become susceptible again and in state $X_i(t)=0$. To be more specific, every node $i$ at time $t$ is infected with probability Pr$[X_i(t)=1]$ and susceptible with probability Pr$[X_i(t)=0]$. At each time $t$, a node can only be in either of these two states, thus Pr$[X_i(t)=1]+$Pr$[X_i(t)=0]=1$. {\color{blue}The state for each individual at time $t$ is independent.} 

Then the following heterogeneous weighted SIS-based network model is obtained from the N-intertwined mean-field approximation (NIMFA)~\cite{van2009virus,devriendt2017unified,hu2018individual}:
{\color{blue}
\begin{equation}
{{\dot p}_i}(t) = - {\gamma _i}{p_i}(t)+(1 - {p_i}(t))\sum\limits_{j = 1}^N {{w_{ij}}(t){\beta _j}{p_j}(t)},\;i = 1,\cdots, N \label{eqn:1}
\end{equation}}
where ${p_i}(t) \in \left[ {0,1} \right]$ denotes the probability of node $i$ being infected at time $t\geq0$. The infection and curing rates $\beta_i\geq0$ and $\gamma_i\geq0$ for each node $i$ in the network are described by two independent Poisson processes; $w_{ij}(t)\in[0,1]$ denotes the weight of edge from node $j$ to node $i$; $N$ is the number of nodes in the network. 

Then (\ref{eqn:1}) can be rewritten in the following compact form: 
\begin{equation}
\dot {p}(t) = (W(t)B - D)p(t) - P(t)W(t)Bp(t), \label{eqn:2}
\end{equation}
where $p(t) = {[{p_1}(t), \cdots ,{p_N}(t)]^T}, W(t) = {[{w_{ij}}(t)]_{N \times N}}, B = diag[{\beta _1}, \cdots ,{\beta _N}], D = diag[{\gamma _1}, \cdots ,{\gamma _N}]$ and $P(t) = diag[{p_1}(t), \cdots ,{p_N}(t)].$ {\color{blue}Here ``$diag$" denotes a diagonal matrix.} 

To be noticed, the network considered here is a directed one and the weights of which are in association with the infected level in the network as it is shown in (\ref{eqn:2}). Define $w_{ii}=0$ for all node $i$ so that self-loop is not considered here.   

\subsection{Problem Formulation}
As it is shown in (\ref{eqn:2}), the infected level can be controlled by the weights adaptation in the network. However, bearing the cost of weights adaptation in mind, a natural dynamic constrained optimization problem is formed as follows:
\begin{equation}
\label{eqn:3}
\mathop {\min }\limits_{w \in W} f = \int_0^T {\{ \sum\limits_{i = 1}^N {{f_i}({p_i}(t))} \} dt} \\
\end{equation}

\begin{eqnarray}
s.t.:g &=& \int_0^T {\{ \sum\limits_{i = 1}^N {\sum\limits_{j = 1}^N {{g_{ij}}({w_{ij}}(t) - w_{ij}^0)} } \} dt}-C\leq0;   \nonumber \\
{{\dot p}_i}(t) &=& (1 - {p_i}(t))\sum\limits_{j = 1}^N {{w_{ij}}(t){\beta _j}{p_j}(t)}  - {\gamma _i}{p_i}(t); \nonumber \\
{p_i}(0) &=& {p_0}(i),0 \leq p_i(t) \leq 1, 0 \leq w_{ij}(t) \leq 1, \nonumber \\
1 &\leq &i\leq N,1 \leq j \leq N. \nonumber
\end{eqnarray} 
where $W$ is the set of all admissible weights; $f_i(p_i(t))$ denotes the infection cost function for each individual $i$, the objective function $f$ denotes the total infection cost for all individuals in the considered period $[0,T]$; ${{g_{ij}}({w_{ij}}(t) - w_{ij}^0)}$ denotes the cost function for weights adaptation and $ w_{ij}^0$ denotes the initial weight at $t=0$, $C$ is a constant that characterizes the maximum cost of weight adaptation, $g$ is the inequality constraint.  

The dynamic constrained optimization problem is interpreted as: how to design the adaptive weights $\{w_{ij}(t)\}$ for all links from $t=0$ to $t=T$ such that the infected level in the network can be suppressed to the maximally extent under given budgets.

\section{Differential Evolution}
\label{section:DE and FR}
Differential evolution (DE) \cite{storn1997differential,wang2018composite} is used as a base optimizer to solve the dynamic constrained optimization problem in (\ref{eqn:3}), which is arguably one of the most powerful stochastic real-parameter optimization algorithms in current use. Different from traditional EAs, DE uses difference of individual trial solutions to explore the objective function landscape. Typically, DE consists of four stages: initialization, mutation,
crossover, and selection.\\

\quad \emph{Step 1: Initialization}\\ 
The ultimate goal of DE is to find a global optimum point in a $D$-dimensional real parameter space $\mathbb{R}^D$. In the beginning, a randomly initiated population $P$ which consists of $NP$ {\color{blue}(size of population)} $D$-dimensional individuals is generated. {\color{blue}$NP$ denotes the size of population where each individual with size $N \times (N-1) \times (T-1)$ denotes a solution that consists of the weighs of every directed link in the spreading network from time $t=1$ to time $T-1$.} The subsequent generations in DE are denoted by $G=1,\cdots,G_{max}$. The population at generation $G$ is denoted as follows:
\begin{displaymath}
P_G = \left\{ {{{\vec x}_{1,G}}, \cdots ,{{\vec x}_{NP,G}}} \right\},
\end{displaymath}
where ${\vec x}_{i,G}$ is the $i$th target vector in current generation:
\begin{displaymath}
{{\vec x}_{i,G}} = [{x_{1,i,G}}, \cdots ,{x_{D,i,G}}].
\end{displaymath} 
Initially ($G=0$), the population should be uniformly randomized within the search space constrained by the maximum and minimum bounds: ${{\vec x}_{\min }} = \left[ {{x_{1,\min }}, \cdots ,{x_{D,\min }}} \right]$ and ${{\vec x}_{\max }} = \left[ {{x_{1,\max }}, \cdots ,{x_{D,\max }}} \right]$. Hence the initialization is formulated as follows:
\begin{displaymath}
{x_{j,i,0}} = {x_{j,\min }} + ran{d_{i,j}}\left[ {0,1} \right] \cdot ({x_{j,\max }} - {x_{j,\min }}),
\end{displaymath} 
where $ran{d_{i,j}}\left[ {0,1} \right]$ is a uniformly distributed number between $0$ and $1$. \\

\emph{Step 2: Mutation}\\  
The mutation operator aims to create a mutant vector for each target vector through utilizing the differential information of pairwise individuals. The following mutation operator is adopted in this paper.

DE/current-to-best/1:
\begin{displaymath}
{\vec v _{i,G}} = {\vec x _{i,G}} + F({\vec x _{best,G}} - {\vec x _{i,G}}) + F({\vec x _{r_1^i,G}} - {\vec x _{r_2^i,G}}),
\end{displaymath}
where $i=1,\cdots,NP$, ${\vec v _{i,G}}=[{v_{1,i,G}}, \cdots ,{v_{D,i,G}}]$ is the mutant vector, $r^i_1$ and $r^i_2$ are mutually exclusive integers randomly chosen from $[1,NP]\backslash i$, ${\vec x _{best,G}}$ is the best individual in the current population, and the scaling factor $F$ is a positive control parameter for scaling the difference vectors.\\   

\emph{Step 3: Crossover}\\  
The crossover operator is employed to enhance the diversity of the population. The trial vector ${\vec u _{i,G}}=[{u_{1,i,G}}, \cdots ,{u_{D,i,G}}]$ is formed by exchanging components between the target vector ${\vec x}_{i,G}$ and the mutant vector ${\vec v}_{i,G}$. The following binomial crossover is utilized:

\begin{numcases}{u_{j,i,G}=}
v_{j,i,G}, & if $(rand_{i,j}[0,1] \leq {Cr}\;\text{or}\;j = {j_{rand}})$ \nonumber \\
x_{j,i,G}, & otherwise.\nonumber
\end{numcases}
where $i=1,\cdots,NP$, $j=1,\cdots,D$, $j_{rand}$ is a randomly chosen index from $[1,D]$. $Cr$ is the crossover rate.\\ 

\emph{Step 4: Selection}\\  
In the selection step, the target vector ${\vec x}_{i,G}$ is compared with the trial vector ${\vec u _{i,G}}$, the better one can be survived in the next generation
\begin{numcases}{\vec x _{i,G + 1} = }
\vec u _{i,G}, & if $f({\vec u _{i,G}}) < f({\vec x _{i,G}})$ \nonumber \\
\vec x _{i,G}, & otherwise.\nonumber
\end{numcases}

\section{Methodology}
\label{section:methodology}
\subsection{Encoding Mechanism}
As described above, the dimension of the decision space $D$ is related to the number of links in the network and the time length $T$. To be more specific, for a directed network with $N$ nodes, we have 
\begin{equation}
\label{eqn:4}
D=N\times(N-1)\times (T-1). 
\end{equation}   

\begin{figure*}[!t]
	\centering
	\includegraphics[width=6in]{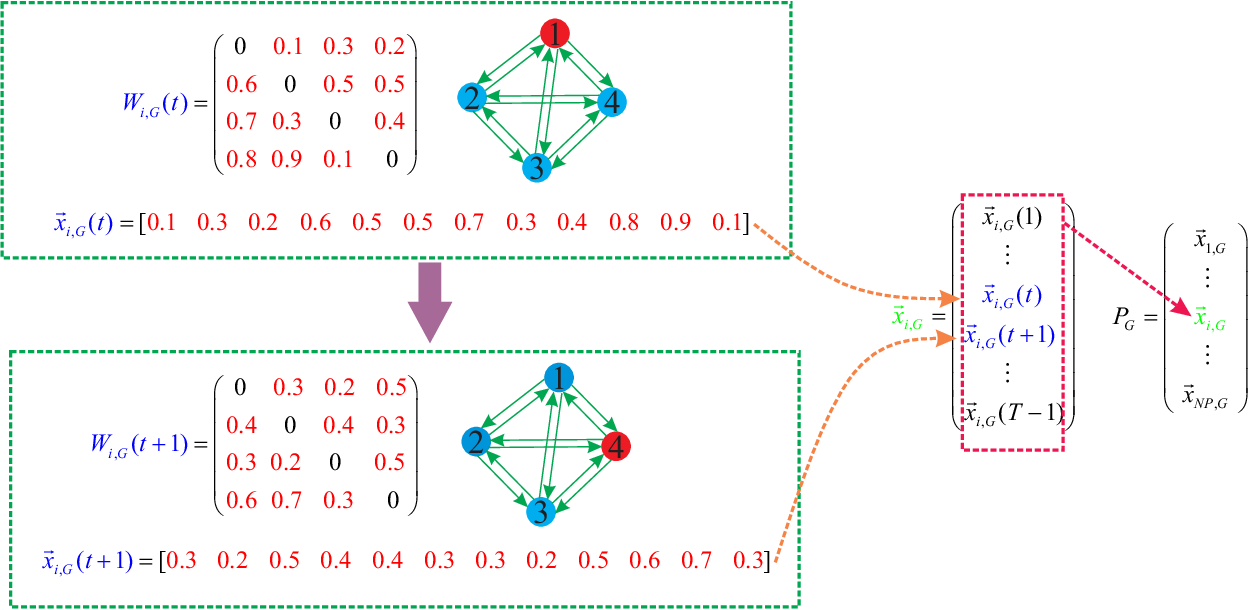}
	\caption{Schematic of the encoding mechanism.}
	\label{fig_encode}
\end{figure*}

To better illustrate the encoding mechanism, a schematic graph is presented in Figure~\ref{fig_encode}. The network consists of $N=4$ nodes, where blue and red nodes represent susceptible and infected individuals, respectively. $\vec{x}_{i,G}(t)$ denotes the $i$th target vector in the $G$-th generation, with a dimension of $N\times(N-1)$. Each element of $\vec{x}_{i,G}(t)$ corresponds to a specific element of the weight matrix $W_{i,G}(t)$ at time $t$ in the manner presented in Figure~\ref{fig_encode}. Bearing the infection cost and the weights adaption constraint in mind, the weight matrix $W_{i,G}(t+1)$ at time $t+1$ is evolved adaptively to balance the objective function and constraint. Considering the time length $T$, the $i$th target vector at $G$-th generation is formulated as:

\begin{displaymath}
\vec{x}_{i,G}=[\vec{x}_{i,G}(1),\cdots,\vec{x}_{i,G}(t),\vec{x}_{i,G}(t+1),\cdots,\vec{x}_{i,G}(T-1)].
\end{displaymath} 

To be noticed, the health status of all individuals in the network is varying from time to time. For example, as it is shown in Figure~\ref{fig_encode}, node $1$ is infected at time $t$ while it is cured and become susceptible at time $t+1$. Therefore, this is a dynamic optimization problem considering the interactions between the epidemic spreading process and weights adaptation.

\begin{remark}
	With this encoding mechanism, one inevitable problem is that the dimension of the decision space increases exponentially with the size of the network $N$ as it is shown in (\ref{eqn:4}). Therefore, this constrained optimization problem (\ref{eqn:3}) may suffers from the ``curse of dimensionality", which implies that most of the EAs' performance deteriorates rapidly as the increasing of the dimensionality of the search space.       
\end{remark} 

\subsection{Differential Evolution with Neighborhood Search (NSDE)}
\label{section:NSDE}
As a variant of classical DE introduced in Section \ref{section:DE and FR}, NSDE~\cite{yang2007making} is effective in escaping from local optima when searching in circumstances without knowing the preferred step size. The main difference between NSDE and DE is that the neighborhood search (NS) strategy is utilized, which is a typical technique in evolutionary programming (EP)~\cite{yao1999evolutionary}. To be more specific, the scaling factor $F$ in classical DE is replaced in the following manner:

\begin{numcases}{F_{i}=}
N_{i}(0.5,0.5), & if $(rand_{i}[0,1] < f_p)$ \nonumber \\
\delta_{i}, & otherwise.\nonumber
\end{numcases}  
where $N_{i}(0.5,0.5)$ is a Gaussian random number with mean 0.5 and standard deviation 0.5, and $\delta_{i}$ is a Cauchy random variable with scale parameter $t=1$. In NSDE, the parameter $f_p$ was set to a constant number 0.5.  

\subsection{$\epsilon$ Constraint-handling Technique}
\label{section:constraint handling}
The $\epsilon$ constraint-handling technique \cite{takahama2010constrained}, which is adopted by the winner of IEEE CEC2010 competition, is utilized to compare two target vectors $\vec{x}_i$ and $\vec{x}_j$. To be more specific, $\vec{x}_i$ is better than $\vec{x}_j$ if the following conditions are satisfied:

\begin{subnumcases}{}\label{eqn:constraint} 
f(\vec x _{i})<f(\vec x _{j}), & if $G({\vec x _{i}}) \leq\epsilon \wedge G({\vec x _{j}}) \leq\epsilon$  \\
f(\vec x _{i})<f(\vec x _{j}), & if $G({\vec x _{i}})=G({\vec x _{j}})$  \\
G(\vec x _{i})<G(\vec x _{j}), & otherwise. 
\end{subnumcases}
where $G(\vec{x})$ denotes {\color{blue}the} degree of constraint violation on the constraint as follows:
\begin{displaymath}
G(\vec{x})=\max(0,g(\vec{x})).
\end{displaymath}

In Eq.~\ref{eqn:constraint}, $\epsilon$ is designed to decrease with the increasing of the generation $G$ as follows~\cite{wang2018improved}:

\begin{numcases}{\epsilon = }
\epsilon_0(1-\frac{G}{G_{max}})^{cp}, & if $G\leq Gc$ \nonumber \\
0, & otherwise.\nonumber
\end{numcases}
\begin{displaymath}
cp=-\frac{\log\epsilon_0+\lambda}{\log(1-\frac{Gc}{G_{max}})},
\end{displaymath}
where $\epsilon_0$ is the maximal degree of constraint violation of the initial population; $Gc$ is a parameter to truncate the value of $\epsilon$; $\lambda$ is set to be $10$ in this paper.

\subsection{A Constrained Cooperative Coevolution ($C^3$) Strategy}
In this subsection, a novel coevolution strategy is developed to solve this dynamic constrained optimization problem motivated by the DECC-G algorithm in \cite{yang2008large}. The core of this coevolution strategy is ``divide-and-conquer". That is, the original $D$-dimensional search space is divided into $N_s$ number of $D_s$-dimensional ones by random grouping strategy, where $N_s\times D_s=D$. 

The constrained cooperative coevolution ($C^3$) framework for this problem can be summarized as follows:

(1) Set $i=1$ to start a new \emph{cycle}. 

(2) Decompose an original $D$-dimensional target vector into $N_s$ low-dimensional subcomponents with dimension $D_s$ randomly, i.e. $D=N_s\times D_s$. Here ``randomly" indicates that each dimension in the original target vector has the same probability to be assigned into any of the $N_s$ subcomponents. 

(3) Optimize the $j$th subcomponent with NSDE and $\epsilon$ constraint-handling technique introduced in Section \ref{section:NSDE} and Section \ref{section:constraint handling} for a predefined number of fitness evaluations (FEs). 

(4) if $j<N_s$ then $j++$ and go to Step 3. 

(5) Stop if halting criteria are satisfied; otherwise go to Step (1) for the next \emph{cycle}.

\begin{remark}
	The probability of $C^3$ strategy to assign two interacting variables $x_i$ and $x_j$ into a single subcomponent for at least $k$ cycles is:
	\begin{displaymath}
	{{P}_k} = \sum\limits_{l = k}^K {{K \choose l}{{\left( {\frac{1}{{{N_s}}}} \right)}^l}{{\left( {1 - \frac{1}{{{N_s}}}} \right)}^{K - l}}} 
	\end{displaymath} 
	where $K$ is the total number of cycles and $N_s$ is the number of subcomponents.
	 
	{\color{blue}
	In each separate cycle, the probability to assign two interacting variables $x_i$ and $x_j$ into a single subcomponent is 
	\begin{displaymath}
	p=\frac{{N_s \choose 1}}{N_s^2}=\frac{1}{N_s}.
	\end{displaymath}
	Let $p_k$ denotes the probability to assign $x_i$ and $x_j$ into a single subcomponent for exactly $k$ cycles. Obviously, $p_k$ satisfies the binomial distribution, so:
	\begin{displaymath}
	p_k={{K \choose l}{p^l}{{\left( {1 - p} \right)}^{K - l}}}={{K \choose l}{{\left( {\frac{1}{{{N_s}}}} \right)}^l}{{\left( {1 - \frac{1}{{{N_s}}}} \right)}^{K - l}}}.
	\end{displaymath} 
	
	Thus, 
	\begin{displaymath}
	{{P}_k} = \sum\limits_{l = k}^K{p_k}= \sum\limits_{l = k}^K {{K \choose l}{{\left( {\frac{1}{{{N_s}}}} \right)}^l}{{\left( {1 - \frac{1}{{{N_s}}}} \right)}^{K - l}}}. 
	\end{displaymath}}	
\end{remark}

Given network size $N=20$ and $T=10$, then $D=N\times(N-1)\times(T-1)=3420$. Select $D_s=N\times(N-1)=380$, then the number of subcomponents $N_s=9$. When the number of cycles $K=50$, we have:
\begin{eqnarray}
{P_1} &=& 1 - {(1 - \frac{1}{9})^{50}} = 0.9972. \nonumber \\ 
{P_2} &=& {P_1} - {50 \choose 1}\times\frac{1}{9} \times {(1 - \frac{1}{9})^{49}}=0.9799. \nonumber
\end{eqnarray}   
These results demonstrate that the $C^3$ strategy has relatively high probabilities to optimize interacting variables in a single subcomponent for at least one or two cycles. 

\subsection{NSDE under the $C^3$ framework}
{\color{blue}Considering} NSDE as the base optimizer for the subcomponents, it is straightforward to obtain the NSDE with constrained cooperative coevolution algorithm, denoted as NSDE-$C^3$. The pesudocode of NSDE-$C^3$ is given in Algorithm~\ref{alg:NSDEC3}. 

\begin{algorithm}[!h]
	/*{\it Initialization}*/ \\
	$pop(1:NP, 1:D) \leftarrow  rand(popsize, D)$\;
	$(best,bestval)\leftarrow evaluate(pop)$\;
	\For{$i=1:cycles$}
	{
		$index(1:D)\leftarrow randperm(D)$\;
		\For{$j=1:N_s$}
		{
			$k\leftarrow (j-1)\times D_s+1$\;
			$l\leftarrow j \times D_s$\;
			$subpop(1:NP, 1:D_s) \leftarrow  pop(:, index(k:l))$\;
			/*{\it Use sub-optimizer}*/ \\
			$subpop \leftarrow NSDE(best, subpop, FEs)$\;
			$ pop(:, index(k:l))\leftarrow subpop(1:NP, 1:D_s) $\;
			$(best,bestval)\leftarrow evaluate(pop)$\;
		}	  	
	}
	\Return{$pop(best,:), bestval$}    	
	\caption{NSDE-$C^3$}\label{alg:NSDEC3}
\end{algorithm}	

\section{Numerical Experiments}
\label{section:Numerical Experiments}
To validate the effectiveness of the proposed method, a synthetic network is obtained as the underlying network. When solving the dynamic constrained optimization problem in (\ref{eqn:3}), each optimization algorithm is used for 25 independent runs. 

\subsection{Datasets}
For validation, one of the most famous synthetic complex network-the B\' arabasi-Albert (BA) network \cite{barabasi1999emergence} is used. When constructing the BA network, initially $m_0=5$ fully connected nodes are placed in the network. A new node is connected to $m=m_0=5$ existing nodes with probability proportional to the degree of them at each step \cite{feng2016PA}. A total number of $N=20$ nodes is fixed for the BA network in the simulations, originally this BA network is a bidirectional network.    

\begin{figure}[!b]
	\centering
	\includegraphics[width=1.8in]{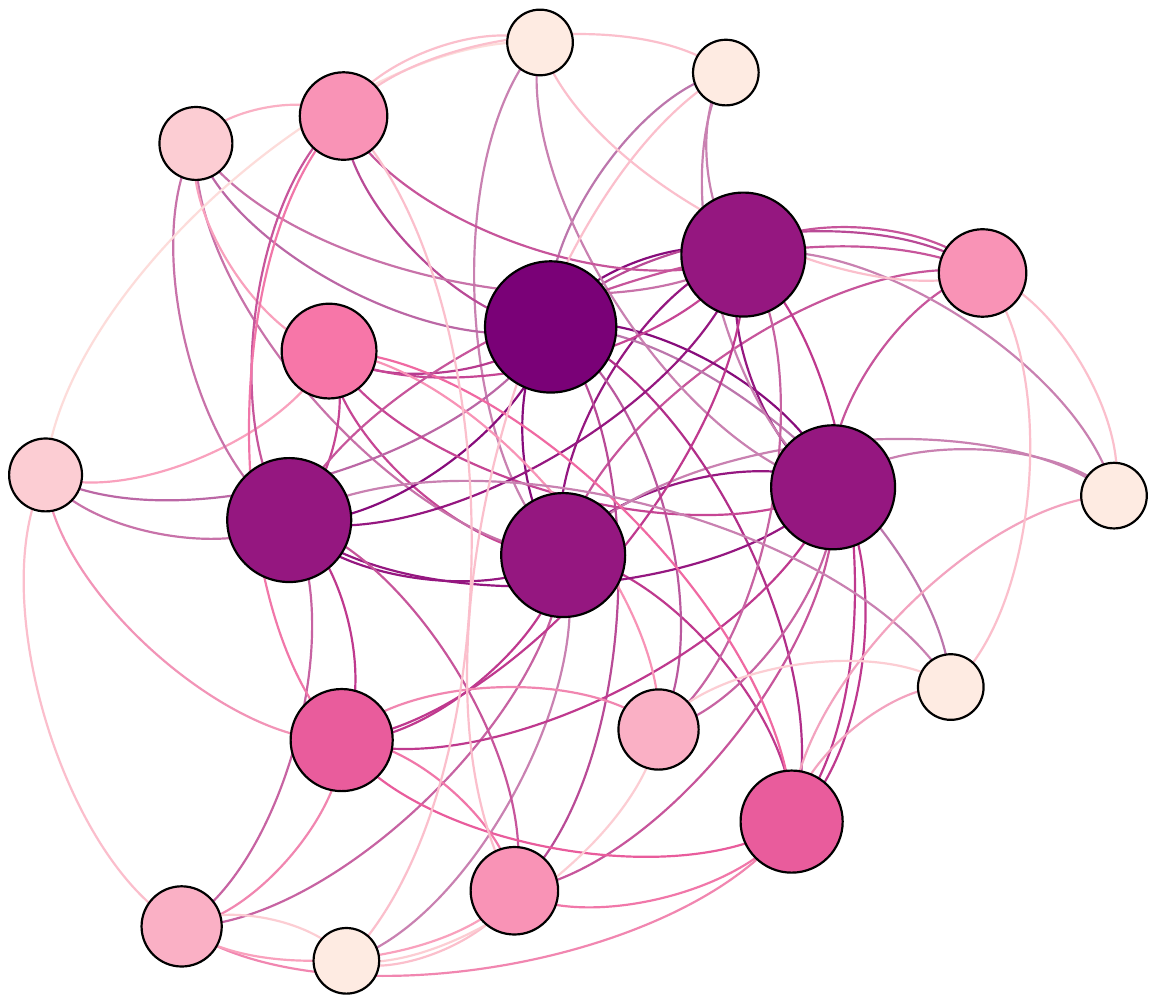}
	\caption{B\'arabasi-Albert network}
	\label{fig_BA}	
\end{figure}

The topologies of the BA network is presented in Figure~\ref{fig_BA}, where circles represent nodes in the network and their sizes are proportional to degrees.   

Table~\ref{tab:topology} below summarizes the topological features of the network, where $N, <k>, <C>, d$ denote the total number of nodes, average degree, average clustering coefficient and density of the network, respectively.  
\begin{table}[!h] 
	\centering
	\caption{Topological features of the analyzed network.}
	\label{tab:topology}
	\begin{tabular}{l|cccl} 
		\toprule 
		Network name& $N$& $<k>$ & $<C>$& $d$ \\ 
		\midrule 
		B\' arabasi-Albert & 20 & 8.5 & 0.519 & 0.447 \\ 
		\bottomrule 
	\end{tabular} 
\end{table}

%

%

\subsection{Parameters Selection}
\label{section:Parameters Selection}

For all experiments, the objective function for individual $i$ at time $t$ adopts the form: 
\begin{displaymath}
f_i(p_i(t))=\sqrt{p_i(t)}.
\end{displaymath}   
Meanwhile, the cost function for weights adaptation for a pair of individuals $i$ and $j$ is:
\begin{displaymath}
g_{ij}(w_{ij}(t))=(w_{ij}(t)-w^0_{ij})^2.
\end{displaymath}
The value of the constraint $C$ is selected as 700.  

{\color{blue}The objective function $f_i(p_i(t))$ and the cost function $g_{ij}(w_{ij}(t))$ adopt the current form for simple illustration. They represent one possible situation, there exists other sets of $f_i(p_i(t))$ and $g_{ij}(w_{ij}(t))$ in real-world scenarios. Since the weights adaptation optimization problem is still an open-problem and there lacks standard definition of both the objective function and the cost function, we adopted the current form similar to the way that was done in \cite{feng2018CNSNS} and \cite{hu2018individual}. 
	
	As for the value of the constraint $C$, it was selected after the spreading network and epidemic parameters have been set.}

Based on the underlying network introduced in the previous subsection, some numerical simulations are conducted to illustrate the effectiveness of the proposed method (NSDE-$C^3$). 

The epidemic parameters used in the simulations are presented in Table~\ref{tab:parameters}, where $p_i(0), \beta_i, \gamma_i, T$ denote the initial infection state, infection rate, curing rate, and terminal time, respectively. These parameters are carefully chosen to make sure that the infected level is relatively high when there is no weights adaptation, hence the effects of weights adaptation on epidemic spreading can be easily observed. 

{\color{blue}
\begin{remark}\label{rem:threshold}
	According to Theorem 2 in \cite{khanafer2014stability}, assume that $p(0)\ne 0$, then the metastable state $p^\star$ is globally asymptotically stable when 
	\begin{displaymath}
	\lambda_{max}(AB-D)>0
	\end{displaymath}
	where ${\lambda _{\max }}(AB-D)$ is the largest eigenvalue-the spectral radius of the matrix $AB-D$. Referring to the epidemic parameters in Table~\ref{tab:parameters}, it is easily obtained that $\lambda_{max}(AB-D)=3.5167>0$ for the constructed BA network, hence the infected level is relatively high without weights adaptation.    
\end{remark}
}

Regarding to the algorithm parameters, the population size $NP$, the number of FEs and the crossover rate for both NSDE and NSDE-$C^3$ algorithms are the same. For NSDE-$C^3$ strategy, the dimension of the subcomponents $D_s$ is selected as $N(N-1)$.  

\begin{table}[!t] 
	\centering
	\caption{Epidemic and algorithm parameters used in the simulations.}
	\label{tab:parameters}
	\begin{tabular}{l|ccccccl} 
		\toprule 
		Network name& $p_i(0)$& $\beta_i$ & $\gamma_i$ & $T$ & $NP$ & FEs & $Cr$ \\ 
		\midrule 
		B\' arabasi-Albert& 0.153 & 0.4 & 0.3 & 10 & 350 & 6.30e+06 &0.9  \\ 
		\bottomrule 
	\end{tabular} 
\end{table}

\subsection{Results}
\begin{figure}[!t]
	\centering
	\subfigure[Fitness value]{
		\includegraphics[width=2.5in]{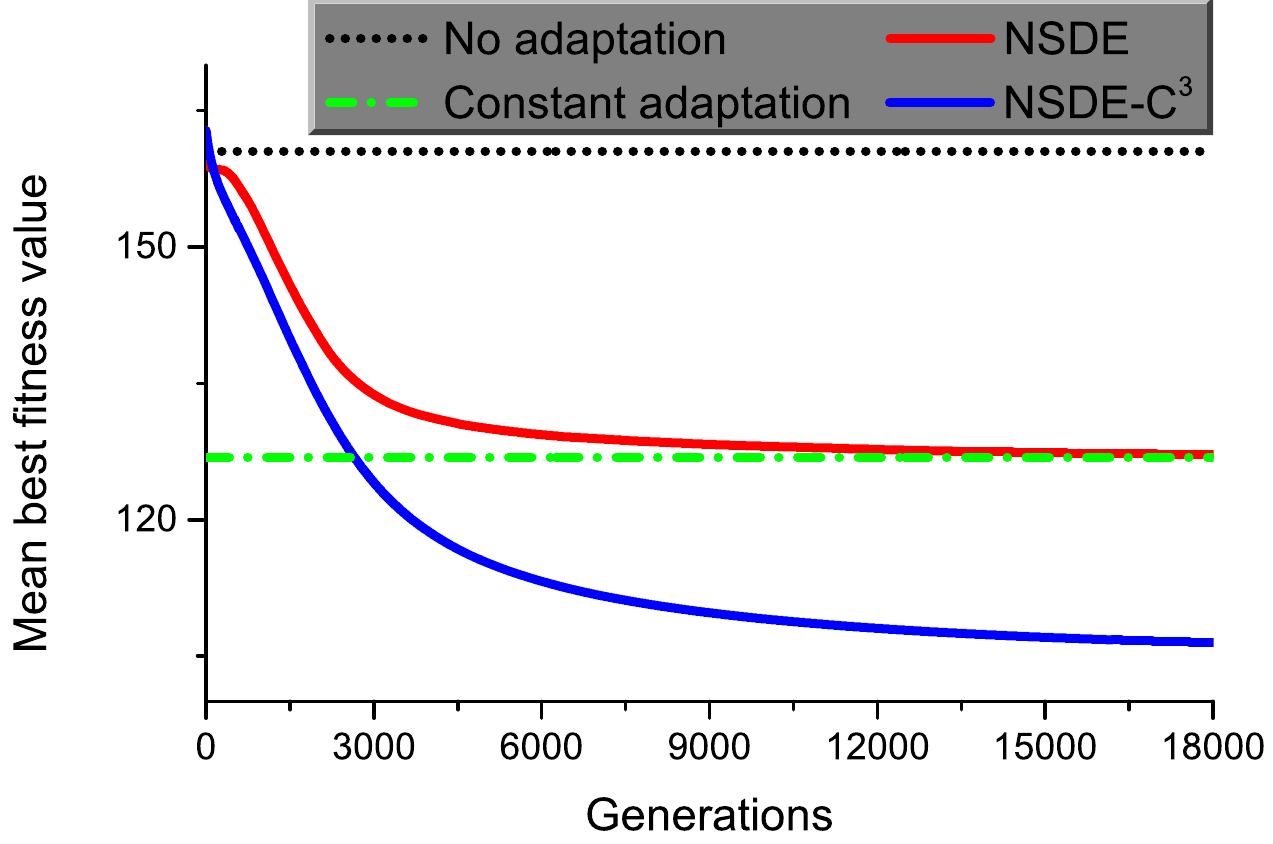}
		\label{fig_BA_f}		
	}    
	\subfigure[Constraint violation]{
		\includegraphics[width=2.5in]{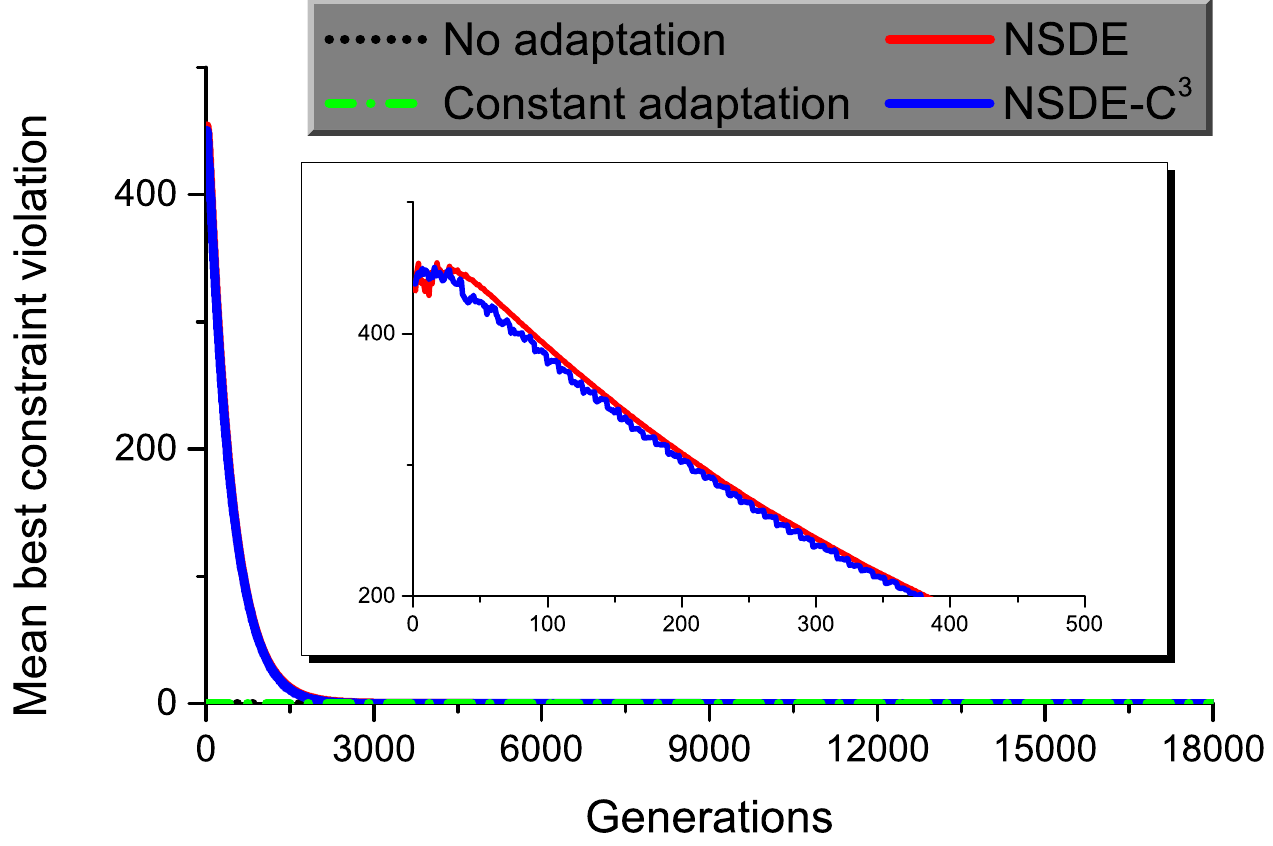}
		\label{fig_BA_g}
	}
	\subfigure[Epidemic evolution]{
		\includegraphics[width=2.5in]{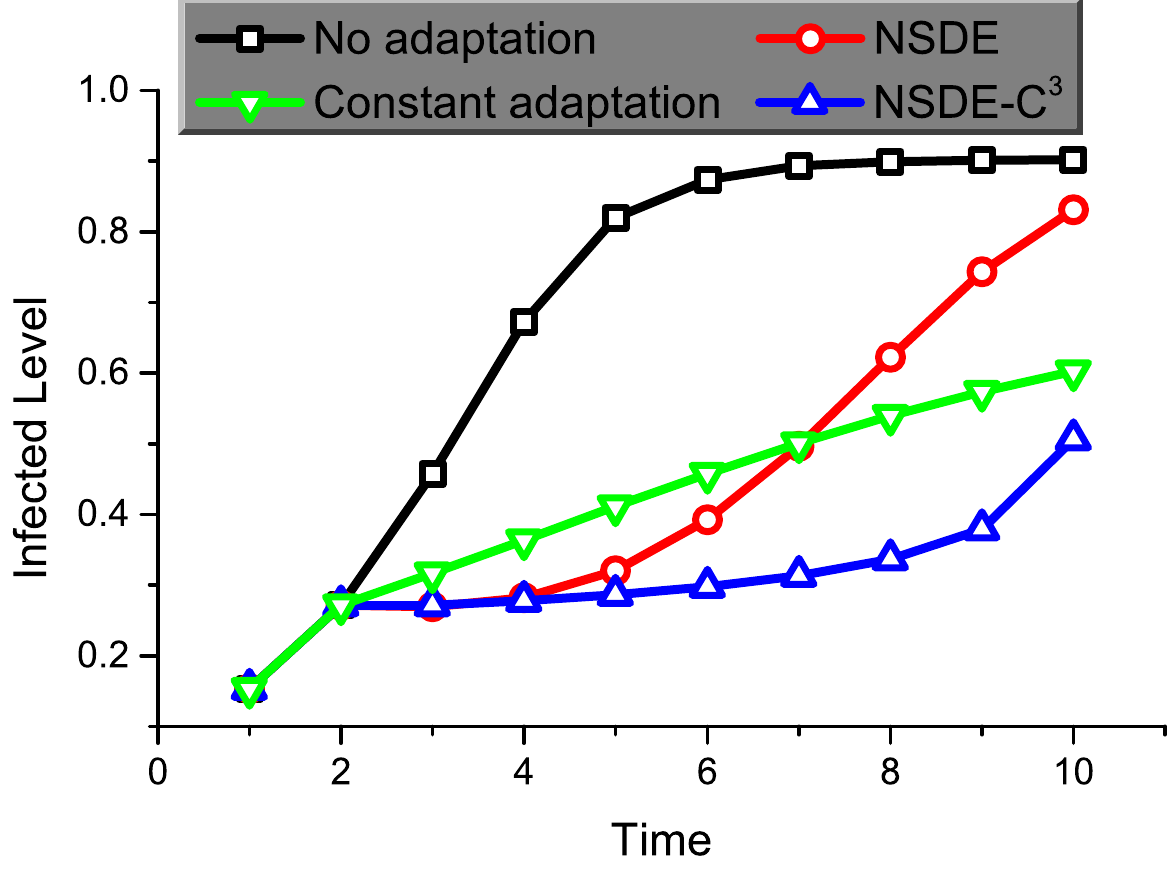}
		\label{fig_BA_I}		
	}    
	\subfigure[Weights adapation]{
		\includegraphics[width=2.5in]{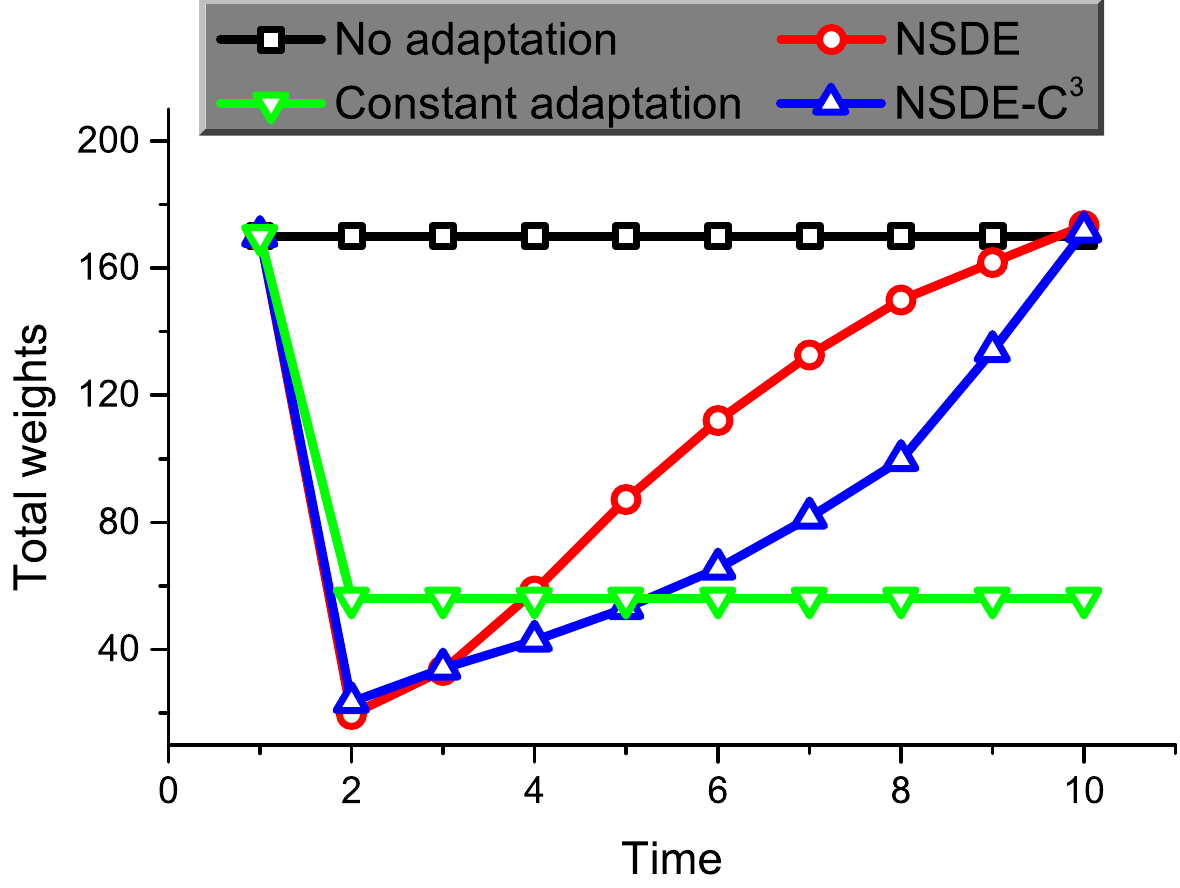}
		\label{fig_BA_W}
	}
	\caption{Simulation results on BA network for 25 independent runs}
	
\end{figure}

For B\' arabasi-Albert network, the simulation results are presented in Figure \ref{fig_BA_f}, \ref{fig_BA_g}, \ref{fig_BA_I} and \ref{fig_BA_W}. Both the proposed NSDE-$C^3$ and the existing NSDE method are used. Meanwhile, two control groups are added for comparison, i.e. ``No adaptation'' and ``constant adaptation'' strategy. The formulation of these two strategies at time $t$ are as follows:
\begin{eqnarray}
\text{No adaptation:}\; w_{ij}(t)&=&w_{ij}^0; \nonumber \\
\text{Constant adaptation:}\; w_{ij}(t)&=&c\times w_{ij}^0,\nonumber \text{where} 
\end{eqnarray}
\begin{displaymath}
g=\int_0^T {\{ \sum\limits_{i = 1}^N {\sum\limits_{j = 1}^N {{g_{ij}}(w_{ij}(t) - w_{ij}^0)} } \} dt}-C=0.
\end{displaymath}
In this manner, these two strategies are employed as the baselines for comparison. ``No adaptation'' means that the weights remain unchanged in the considered time period, while ``Constant adaptation'' refers to a ``discount'' on the original weights $w_{ij}^0$ with the budget $C$ fully used. For the epidemic parameters in Table \ref{tab:parameters} and $C=700$, the constant adaptation ratio can be calculated as $c=0.33$.       

Figure~\ref{fig_BA_f} and \ref{fig_BA_g} illustrate the evolution process of the mean best fitness value and constraint violation with respect to generations over 25 independent runs, respectively. For the ``No adaptation'' and ``Constant adaptation'' strategy, the solutions are determined initially. Hence the value of the objective function and constraint violation for these two strategies remain unchanged. For the ``No adaptation'' and ``Constant adaptation'' strategies, the constraint violation remains zero. Combing Figure~\ref{fig_BA_f} and \ref{fig_BA_g}, it is evident that the proposed NSDE-$C^3$ outperforms NSDE on BA network. Moreover, NSDE-$C^3$ outperforms both the ``No adaptation'' and ``Constant adaptation'' strategies. 

\begin{remark}
	One interesting phenomenon to be noticed in Figure~\ref{fig_BA_g} is that the evolution of constraint violation for NSDE-$C^3$ almost coincides with that of NSDE. However, the evolution of the mean best fitness value for these two algorithms are quite different. This is due to the use of $\epsilon$ constraint-handling technique in Section \ref{section:constraint handling}, which puts more emphasis on the constraint violation than the fitness value for earlier generations.     	
\end{remark}

Figure \ref{fig_BA_I} and \ref{fig_BA_W} demonstrate the evolution process of two main indices, i.e. infected level and total weights over time, which characterize the epidemic spreading and weights adaptation level, respectively. The definition of theses two indices are as follows:
\begin{eqnarray}
I(t)&=&\mathbb{E}(p_i(t))=\frac{1}{{N}}\sum\limits_{i = 1}^N {{p_i}(t)}; \nonumber\\
W(t)&=&\sum\limits_{i = 1}^N {\sum\limits_{j = 1, j\ne i}^N {{w_{ij}}(t)} }. \nonumber
\end{eqnarray} 
The physical meaning of the infected level $I(t)$ and the total weights $W(t)$ are the expectation of infection probability and total number of weights, respectively. In this manner, $I(t)$ can be used as an index which reflects the current infected level among the whole population of individuals in the network. Similarly, $W(t)$ reflects the weights variation of the network concerned. For infected level, ``No adaptation'' strategy achieves a relatively high level, which is used as a baseline for comparison as it is demonstrated in Section \ref{section:Parameters Selection} and Remark \ref{rem:threshold}. However, the ``Constant adaptation" strategy achieves a lower infected level than the ``No adaptation'' strategy, which is mainly due to that the ``Constant adaptation" strategy makes full use of the budget as it is shown in Figure \ref{fig_BA_W}. Referring to the total weights $W(t)$ variation in Figure \ref{fig_BA_W}, it is obvious that $W(t)$ remains unchanged at every time $t$ for the ``No adaptation'' strategy. For the ``Constant adaptation'' strategy, the value of the total weights $W$ decreases to a lower value and keeps unchanged for the whole time period. However, for both the NSDE and NSDE-$C^3$ strategies, the value of total weights $W(t)$ experience a {\color{blue}decreasing} process first and restore to the initial value of time $t=1$, which coincide with the results in \cite{hu2018individual}. The restoring process of $W(t)$ for NSDE-$C^3$ is more moderate than that of NSDE, hence the infected level $I(t)$ is lower at most time. One interesting observation is that the value of the total weights for NSDE-$C^3$ drops to a lower value than that of the ``Constant adaptation" strategy, and restores to the initial value at time $T$, the infected levels indicate that this kind of ``dynamically adaptation'' is more effective in controlling the epidemic spreading scale.       

In addition, {\color{blue}to} test statistical significance, the multi-problem Wilcoxon's test \cite{wang2018improved} are implemented to compare these methods, the results are shown in Table \ref{tab:ranksum}. Once more, the results indicate that the proposed NSDE-$C^3$ is superior to the other competitors.

\begin{table}[!h] 
	\centering
	\caption{Wilcoxon's test on BA network for 25 independent runs.}
	\label{tab:ranksum}
	\begin{tabular}{l|cccccl} 
		\toprule 
		Algorithm& Mean OFV $\pm$ Std Dev  & $p$-value \\ 
		\midrule 
		NSDE-$C^3$& \bf{106.4530} $\pm$ 0.1506  & -  \\ 
		NSDE&  127.1566 $\pm$ 1.3425  & 1.4157e-09   \\ 
		No adaptation& 160.5280 $\pm$ 0.00 &  9.7285e-11  \\
		Constant adaptation& 126.8617 $\pm$ 0.00 &  9.7285e-11  \\
		\bottomrule 
	\end{tabular} 
\end{table}

{\color{blue}
\begin{remark}
	To be noticed, the SIS model has been analytically and exactly solved in~\cite{shang2012lie,shang2017lie}, where the infection and recovery rate are time-varying. However, the NIMFA model is employed as the epidemic spreading model for its simplicity. In the future, we may extend our results with the help of the Lie algebra method.     
\end{remark}}

\section{Conclusion}     
\label{section:conclusion}
In this paper, a dynamic constrained optimization problem of weights adaptation for heterogeneous
epidemic spreading networks is formulated. Combining constrained cooperative coevolution ($C^3$) strategy with NSDE, a novel NSDE-$C^3$ algorithm is tailored for this problem. Numerical experiments on a BA network showed the effectiveness of this algorithm. In future research, how to expand this algorithm for on-line implementation is an interesting yet challenging topic. {\color{blue}Moreover, the extension to the Mixed SI(R) epidemic dynamics in random graphs with general degree distributions and also the feedback control optimization are also our interests.}  

{\color{blue}
Some insights on the capability of our approach are provided as follows:

\begin{itemize}
	\item First, in our problem setting, the dimension of a solution to be optimized increases exponentially with the increasing of the network size $N$, which is obvious. And the curse of dimensionality problem is inevitable under such situations. However, even in the evolutionary computation community, state-of-the-art methods to deal with such large scale optimization problem is questionable. (The dimension of a solution in~\cite{yang2008large,duan2019cooperative} is only up to $1000$ while it is $3420$ in the numerical simulations studied in this manuscript.)
	\item The goal of this manuscript is not just to purse as high dimension as possible. More importantly, we focus on introducing evolutionary algorithms into the network optimization problems and shed some light on solving such large-scale optimization problems using the idea of cooperative coevolution.          
\end{itemize}   

In the numerical simulations, the number $N=20$ is selected due to the limitations on the hardware. Moreover, we do consider the capacity of the proposed algorithm seriously and there are still ways to improve. For example, for very large $N$, we may consider dividing the solution of each subcomponent with dimension $N \times (N-1)$ into small sub-subcomponents again and then deal with it in the same manner. 
}

\bibliographystyle{bib/IEEEtran}

\bibliography{bib/IEEEabrv,bib/mybib}

\begin{thebibliography}{10}
\providecommand{\url}[1]{#1}
\csname url@samestyle\endcsname
\providecommand{\newblock}{\relax}
\providecommand{\bibinfo}[2]{#2}
\providecommand{\BIBentrySTDinterwordspacing}{\spaceskip=0pt\relax}
\providecommand{\BIBentryALTinterwordstretchfactor}{4}
\providecommand{\BIBentryALTinterwordspacing}{\spaceskip=\fontdimen2\font plus
\BIBentryALTinterwordstretchfactor\fontdimen3\font minus
  \fontdimen4\font\relax}
\providecommand{\BIBforeignlanguage}[2]{{%
\expandafter\ifx\csname l@#1\endcsname\relax
\typeout{** WARNING: IEEEtran.bst: No hyphenation pattern has been}%
\typeout{** loaded for the language `#1'. Using the pattern for}%
\typeout{** the default language instead.}%
\else
\language=\csname l@#1\endcsname
\fi
#2}}
\providecommand{\BIBdecl}{\relax}
\BIBdecl

\bibitem{pastor2015epidemic}
R.~Pastor-Satorras, C.~Castellano, P.~Van~Mieghem, and A.~Vespignani,
  ``Epidemic processes in complex networks,'' \emph{Reviews of modern physics},
  vol.~87, no.~3, p. 925, 2015.

\bibitem{bernoulli1760essai}
D.~Bernoulli, ``Essai d'une nouvelle analyse de la mortalit{\'e} caus{\'e}e par
  la petite v{\'e}role, et des avantages de l'inoculation pour la
  pr{\'e}venir,'' \emph{Histoire de l'Acad., Roy. Sci.(Paris) avec Mem}, pp.
  1--45, 1760.

\bibitem{feng2014epidemic}
Y.~Feng, Q.~Fan, L.~Ma, and L.~Ding, ``Epidemic spreading on uniform networks
  with two interacting diseases,'' \emph{Physica A: Statistical Mechanics and
  its Applications}, vol. 393, pp. 277--285, 2014.

\bibitem{qu2017sis}
B.~Qu and H.~Wang, ``Sis epidemic spreading with heterogeneous infection
  rates,'' \emph{IEEE Transactions on Network Science and Engineering}, vol.~4,
  no.~3, pp. 177--186, 2017.

\bibitem{feng2016CPB}
Y.~Feng, L.~Ding, Y.-H. Huang, and Z.-H. Guan, ``Epidemic spreading on random
  surfer networks with infected avoidance strategy,'' \emph{Chinese Physics B},
  vol.~25, no.~12, p. 128903, 2016.

\bibitem{huang2016epidemic}
Y.~Huang, L.~Ding, Y.~Feng, and J.~Pan, ``Epidemic spreading in random walkers
  with heterogeneous interaction radius,'' \emph{Journal of Statistical
  Mechanics: Theory and Experiment}, vol. 2016, no.~10, p. 103501, 2016.

\bibitem{nadini2018epidemic}
M.~Nadini, A.~Rizzo, and M.~Porfiri, ``Epidemic spreading in temporal and
  adaptive networks with static backbone,'' \emph{IEEE Transactions on Network
  Science and Engineering}, 2018.

\bibitem{nowzari2016analysis}
C.~Nowzari, V.~M. Preciado, and G.~J. Pappas, ``Analysis and control of
  epidemics: A survey of spreading processes on complex networks,'' \emph{IEEE
  Control Systems}, vol.~36, no.~1, pp. 26--46, 2016.

\bibitem{ghezzi1997pid}
L.~L. Ghezzi and C.~Piccardi, ``Pid control of a chaotic system: An application
  to an epidemiological model,'' \emph{Automatica}, vol.~33, no.~2, pp.
  181--191, 1997.

\bibitem{li2017minimizing}
X.-J. Li, C.~Li, and X.~Li, ``Minimizing social cost of vaccinating network sis
  epidemics,'' \emph{IEEE Transactions on Network Science and Engineering},
  no.~1, pp. 1--1, 2017.

\bibitem{nowzari2017optimal}
C.~Nowzari, V.~M. Preciado, and G.~J. Pappas, ``Optimal resource allocation for
  control of networked epidemic models,'' \emph{IEEE Transactions on Control of
  Network Systems}, vol.~4, no.~2, pp. 159--169, 2017.

\bibitem{chen2017optimal}
H.~Chen, G.~Li, H.~Zhang, and Z.~Hou, ``Optimal allocation of resources for
  suppressing epidemic spreading on networks,'' \emph{Physical Review E},
  vol.~96, no.~1, p. 012321, 2017.

\bibitem{pizzuti2018genetic}
C.~Pizzuti and A.~Socievole, ``A genetic algorithm for finding an optimal
  curing strategy for epidemic spreading in weighted networks,'' in
  \emph{Proceedings of the Genetic and Evolutionary Computation
  Conference}.\hskip 1em plus 0.5em minus 0.4em\relax ACM, 2018, pp. 498--504.

\bibitem{xu2017optimal}
D.~Xu, X.~Xu, Y.~Xie, and C.~Yang, ``Optimal control of an sivrs epidemic
  spreading model with virus variation based on complex networks,''
  \emph{Communications in Nonlinear Science and Numerical Simulation}, vol.~48,
  pp. 200--210, 2017.

\bibitem{shang2013optimal}
Y.~Shang, ``Optimal control strategies for virus spreading in inhomogeneous
  epidemic dynamics,'' \emph{Canadian Mathematical Bulletin}, vol.~56, no.~3,
  pp. 621--629, 2013.

\bibitem{feng2018CNSNS}
Y.~Feng, L.~Ding, and P.~Hu, ``Epidemic spreading on random surfer networks
  with optimal interaction radius,'' \emph{Communications in Nonlinear Science
  and Numerical Simulation}, vol.~56, pp. 344--353, 2018.

\bibitem{hu2018individual}
P.~Hu, L.~Ding, and T.~Hadzibeganovic, ``Individual-based optimal weight
  adaptation for heterogeneous epidemic spreading networks,''
  \emph{Communications in Nonlinear Science and Numerical Simulation}, vol.~63,
  pp. 339--355, 2018.

\bibitem{mcasey2012convergence}
M.~McAsey, L.~Mou, and W.~Han, ``Convergence of the forward-backward sweep
  method in optimal control,'' \emph{Computational Optimization and
  Applications}, vol.~53, no.~1, pp. 207--226, 2012.

\bibitem{back1997handbook}
T.~B{\"a}ck, D.~B. Fogel, and Z.~Michalewicz, \emph{Handbook of evolutionary
  computation}.\hskip 1em plus 0.5em minus 0.4em\relax CRC Press, 1997.

\bibitem{vcrepinvsek2013exploration}
M.~{\v{C}}repin{\v{s}}ek, S.-H. Liu, and M.~Mernik, ``Exploration and
  exploitation in evolutionary algorithms: A survey,'' \emph{ACM Computing
  Surveys (CSUR)}, vol.~45, no.~3, p.~35, 2013.

\bibitem{gong2018community}
M.~Gong, C.~Chen, Y.~Xie, and S.~Wang, ``Community preserving network embedding
  based on memetic algorithm,'' \emph{IEEE Transactions on Emerging Topics in
  Computational Intelligence}, no.~99, pp. 1--11, 2018.

\bibitem{du2019network}
W.~{Du}, W.~{Ying}, P.~{Yang}, X.~{Cao}, G.~{Yan}, K.~{Tang}, and D.~{Wu},
  ``Network-based heterogeneous particle swarm optimization and its application
  in uav communication coverage,'' \emph{IEEE Transactions on Emerging Topics
  in Computational Intelligence}, pp. 1--12, 2019.

\bibitem{wu2018particle}
D.~Wu, N.~Jiang, W.~Du, K.~Tang, and X.~Cao, ``Particle swarm optimization with
  moving particles on scale-free networks,'' \emph{IEEE Transactions on Network
  Science and Engineering}, 2018.

\bibitem{das2011differential}
S.~Das and P.~N. Suganthan, ``Differential evolution: a survey of the
  state-of-the-art,'' \emph{IEEE transactions on evolutionary computation},
  vol.~15, no.~1, pp. 4--31, 2011.

\bibitem{wang2011differential}
Y.~Wang, Z.~Cai, and Q.~Zhang, ``Differential evolution with composite trial
  vector generation strategies and control parameters,'' \emph{IEEE
  Transactions on Evolutionary Computation}, vol.~15, no.~1, pp. 55--66, 2011.

\bibitem{liu2010hybridizing}
H.~Liu, Z.~Cai, and Y.~Wang, ``Hybridizing particle swarm optimization with
  differential evolution for constrained numerical and engineering
  optimization,'' \emph{Applied Soft Computing}, vol.~10, no.~2, pp. 629--640,
  2010.

\bibitem{yang2008large}
Z.~Yang, K.~Tang, and X.~Yao, ``Large scale evolutionary optimization using
  cooperative coevolution,'' \emph{Information Sciences}, vol. 178, no.~15, pp.
  2985--2999, 2008.

\bibitem{yang2007making}
Z.~Yang, X.~Yao, and J.~He, ``Making a difference to differential evolution,''
  in \emph{Advances in metaheuristics for hard optimization}.\hskip 1em plus
  0.5em minus 0.4em\relax Springer, 2007, pp. 397--414.

\bibitem{shang2015global}
Y.~Shang, ``Global stability of disease-free equilibria in a two-group si model
  with feedback control,'' \emph{Nonlinear Anal Model Control}, vol.~20, no.~4,
  pp. 501--508, 2015.

\bibitem{van2009virus}
P.~Van~Mieghem, J.~Omic, and R.~Kooij, ``Virus spread in networks,''
  \emph{IEEE/ACM Transactions on Networking (TON)}, vol.~17, no.~1, pp. 1--14,
  2009.

\bibitem{devriendt2017unified}
K.~Devriendt and P.~Van~Mieghem, ``Unified mean-field framework for
  susceptible-infected-susceptible epidemics on networks, based on graph
  partitioning and the isoperimetric inequality,'' \emph{Physical Review E},
  vol.~96, no.~5, p. 052314, 2017.

\bibitem{storn1997differential}
R.~Storn and K.~Price, ``Differential evolution--a simple and efficient
  heuristic for global optimization over continuous spaces,'' \emph{Journal of
  global optimization}, vol.~11, no.~4, pp. 341--359, 1997.

\bibitem{wang2018composite}
B.-C. Wang, H.-X. Li, J.-P. Li, and Y.~Wang, ``Composite differential evolution
  for constrained evolutionary optimization,'' \emph{IEEE Transactions on
  Systems, Man, and Cybernetics: Systems}, 2018.

\bibitem{yao1999evolutionary}
X.~Yao, Y.~Liu, and G.~Lin, ``Evolutionary programming made faster,''
  \emph{IEEE Transactions on Evolutionary computation}, vol.~3, no.~2, pp.
  82--102, 1999.

\bibitem{takahama2010constrained}
T.~Takahama and S.~Sakai, ``Constrained optimization by the $\varepsilon$
  constrained differential evolution with an archive and gradient-based
  mutation,'' in \emph{Evolutionary Computation (CEC), 2010 IEEE Congress
  on}.\hskip 1em plus 0.5em minus 0.4em\relax IEEE, 2010, pp. 1--9.

\bibitem{wang2018improved}
B.-C. Wang, H.-X. Li, and Y.~Feng, ``An improved teaching-learning-based
  optimization for constrained evolutionary optimization,'' \emph{Information
  Sciences}, vol. 456, pp. 131--144, 2018.

\bibitem{barabasi1999emergence}
A.-L. Barab{\'a}si and R.~Albert, ``Emergence of scaling in random networks,''
  \emph{science}, vol. 286, no. 5439, pp. 509--512, 1999.

\bibitem{feng2016PA}
Y.~Feng, L.~Ding, Y.-H. Huang, and L.~Zhang, ``Epidemic spreading on weighted
  networks with adaptive topology based on infective information,''
  \emph{Physica A: Statistical Mechanics and its Applications}, vol. 463, pp.
  493--502, 2016.

\bibitem{khanafer2014stability}
A.~Khanafer, T.~Ba{\c{s}}ar, and B.~Gharesifard, ``Stability properties of
  infected networks with low curing rates,'' in \emph{2014 American Control
  Conference}.\hskip 1em plus 0.5em minus 0.4em\relax IEEE, 2014, pp.
  3579--3584.

\bibitem{shang2012lie}
Y.~Shang, ``A lie algebra approach to susceptible-infected-susceptible
  epidemics,'' \emph{Electronic Journal of Differential Equations}, vol. 2012,
  no. 233, pp. 1--7, 2012.

\bibitem{shang2017lie}
------, ``Lie algebraic discussion for affinity based information diffusion in
  social networks,'' \emph{Open Physics}, vol.~15, no.~1, pp. 705--711, 2017.

\bibitem{duan2019cooperative}
Q.~Duan, C.~Shao, L.~Qu, Y.~Shi, and B.~Niu, ``When cooperative co-evolution
  meets coordinate descent: Theoretically deeper understandings and practically
  better implementations,'' in \emph{2019 IEEE Congress on Evolutionary
  Computation (CEC)}.\hskip 1em plus 0.5em minus 0.4em\relax IEEE, 2019, pp.
  721--730.

\end{thebibliography}

\end{document}